# A Calculus for Causal Relevance

Blai Bonet
Cognitive Systems Laboratory
Department of Computer Science
University of California, Los Angeles
Los Angeles, CA 90024
bonet@cs.ucla.edu

## Abstract

We present a sound and complete calculus for causal relevance that uses Pearl's functional causal models as semantics. The calculus consists of axioms and rules of inference for reasoning about causal relevance relationships. We extend the set of known axioms for causal relevance with new axioms and rules of inference. The axioms are then divided into different sets for reasoning about specific subclasses of models. These subclasses make up a new decomposition of the class of causal models. At the end, we show how the calculus for causal relevance can be used in the task of identifying causal structure from non-observational data.

## 1 Introduction

Causal relevance is concerned with questions of the form: "Given that (variable) $Z$ is fixed, would changing $X$ alter $Y$?" Formal interest in this notion appears in the works of Suppes [8] and Salmon [7] who attempted to give it probabilistic interpretation. This paper pursues a logical approach, and starts with the work by Galles and Pearl [2], which, similar to the work on graphoids [5, 3], is based on a set of axioms and rules of inference that defines a formal deductive system. In this system, causal relevance is expressed by logical formulas and new relevance sentences can be derived from old ones through rules of inference.

Such deductive system requires an interpretation; that is, a function that maps models to the sets of formulas they "satisfy". The models we focus on are Pearl's functional causal models [6]. Once an interpretation is established, a natural question to ask is whether the system is sound (i.e., if every proven statement is true), and whether the system is complete (i.e., if every true statement is provable). In this paper we answer both questions in affirmative for some classes of models.

To do so, we define a hierarchy of causal models (see Halpern [4]), governed by new axioms and rules of inference. Towards the end of the paper (Section 4), we show how information about causal relevance can be used for identifying the structure of causal models from data.

The rest of the paper is organized as follows. In Section 2, we present Pearl's functional causal models, the formal language of causal relevance, the interpretation of the language, and the classes of causal models. Section 3 presents the axiomatizations, some general results about the calculus as well as the proofs for soundness and completeness. Then, Section 4 shows how this theory can be used to solve identification problems. We conclude with a brief summary.

## 2 Model Theory

A causal model is a set of variables in which each variable may have influence on others. The variables are divided in two groups. The *endogenous* variables for which the model provides a description of the mechanisms that influence them, and the *exogenous* variables that influence only endogenous variables and are taken as given; i.e., the model does not explain them. In this section we provide a formal definition of causal models.

Without loss of generality, we can assume that there is one (big) exogenous variable that takes values in a domain for which there are no restrictions. All the endogenous variables for causal models are taken from a *finite* set $\mathcal{X}$. This set is assumed to be fixed for the rest of the paper and often we use it without mention. Some authors like to say that $\mathcal{X}$ is the *signature* of the theory. It is a technical necessity used to restrict the set of models and the formal language for causal relevance.

### 2.1 Causal Models

A causal model $T$ is a tuple $T = (\mathcal{U}, \{F_Y : Y \in \mathcal{X}\})$ where $\mathcal{U}$ is the domain of the unique exogenous vari-able and $\{F_Y : Y \in \mathcal{X}\}$ is a set of equations called *modifiable structural equations*. Each equation is a function

$$F_Y : \prod_{\substack{X \in \mathcal{X} \\ X \neq Y}} \mathcal{D}_X \times \mathcal{U} \longrightarrow \mathcal{D}_Y,$$

where $\mathcal{D}_X$ is the domain of variable $X$, that describes how the variable $Y$ attains its value.

Given a subset $X \subseteq \mathcal{X}$ of endogenous variables, we define the causal (sub)model $[x]T$, that results from the *intervention* of setting the value of $X$ to $x$, as $[x]T = (\mathcal{U}, \{F'_Y : Y \in \mathcal{X}\})$ where $F'_Y = F_Y$ for $Y \notin X$ and $F'_Y = x \downarrow_Y$ for $Y \in X$. This intervention is denoted by the operator $do(X = x)$ that maps causal models into causal models by $T \rightsquigarrow [x]T$ (see [6]). A causal model defines a system of equations once the exogenous variable had been set. In general, the system may have zero, one or multiple solutions. We will only consider causal models $T$ such that for each subset $X \subseteq \mathcal{X}$ and value $x$ for $X$, the submodel $[x]T$ has a unique solution for all $u \in \mathcal{U}$. This class of models was denoted by $\mathcal{T}_{\text{uniq}}$ in [4] (note that $\mathcal{T}_{\text{uniq}}$ depends on $\mathcal{X}$).

The *potential response* for variable $Y$ and $u \in \mathcal{U}$ under intervention $do(X = x)$ is defined as the solution $Y$ in model $[x]T$ when the exogenous variable has value $u$ (denoted by the function $u \rightsquigarrow [x]Y(u)$). The main difference between causal models and Bayesian networks is that the former incorporate a semantics for all interventions $do(X = x)$.

A *counterfactual* is a formula built from atomic terms of the form $[x]Y(u)$ (see Pearl [6]). A logical calculus for reasoning about counterfactuals that is sound and complete with respect to $\mathcal{T}_{\text{uniq}}$ was given in [4]. One result from the counterfactual calculus that is important to us is the following.

**Lemma 1 (Composition; [2])** *Let $W, X, Y$ be variables in a causal model $T$, $x$ a valuation for $X$, and $u$ a valuation for the exogenous variable. If $[x]W(u) = w$ and $[x]Y(u) = y$, then $[x,w]Y(u) = y$.*

Composition can be briefly summarized by the equation $[x]Y(u) = [x, [x]W(u)]Y(u)$ which is valid for all variables $Y, W$, all valuations $x$ of $X$, and all $u \in \mathcal{U}$. As a remark, note that composition is not about nested interventions, it is just a simple intervention that uses a *previously recorded* value from another intervention.

### 2.2 Language

We talk about causal relevance using a simple logical language built from atomic formulas and the standard boolean connectives. The language, that also depends on $\mathcal{X}$, is denoted by $\mathcal{L}$. Each atomic formula, or just atom, of $\mathcal{L}$ is of the form $(X \nrightarrow Y | Z)$ where $X, Y$ and $Z$ are *disjoint* subsets of $\mathcal{X}$, and $X, Y$ are non-empty. Intuitively, an atom $(X \nrightarrow Y | Z)$ is read as *variable $X$ has no influence on variable $Y$ once variable $Z$ had been fixed*. The language is inductively defined by *(i)* every atom is in $\mathcal{L}$, *(ii)* if $\varphi \in \mathcal{L}$, then $\neg \varphi \in \mathcal{L}$, and *(iii)* if $\varphi, \psi \in \mathcal{L}$, then $(\varphi \& \psi), (\varphi \vee \psi), (\varphi \Rightarrow \psi) \in \mathcal{L}$.

As is common, a *literal* is an atom or the negation of an atom. It is positive if it is an atom, otherwise it is negative. We will use capital Greek letters to denote sets of formulas and lower case Greek letters to denote formulas. If $\Gamma$ is a set of formulas, the set of positive (resp. negative) literals in $\Gamma$ is denoted by $\text{Lit}^+ \Gamma$ (resp. $\text{Lit}^- \Gamma$).

### 2.3 Interpretations

An interpretation assigns to each model the set of formulas that it satisfies. We begin with the interpretation for atoms and then extend it to all formulas.

**Definition 1** *A causal model $T \in \mathcal{T}_{\text{uniq}}$ satisfies the atom $(X \nrightarrow Y | Z)$ if and only if*

$$[x, z, w]Y(\cdot) = [x', z, w]Y(\cdot) \quad (1)$$

*for all $W$ disjoint of $X, Y, Z$, and all valuations $x, x', z, w$ for the variables $X, Z$ and $W$.*

Note that Eq.(1) is an equality of functions so it must hold for all $u \in \mathcal{U}$. We write $T \models \varphi$ when $T$ satisfies formula $\varphi$, and $T \models \Gamma$ when $T$ satisfies all formulas in $\Gamma$. The interpretation is extended to all formulas by *(i)* $T \models \neg \varphi$ iff $T \not\models \varphi$, *(ii)* $T \models (\varphi \& \psi)$ iff $T \models \varphi$ and $T \models \psi$, *(iii)* $T \models (\varphi \vee \psi)$ iff $T \models \varphi$ or $T \models \psi$, and *(iv)* $T \not\models (\varphi \Rightarrow \psi)$ iff $T \models \varphi$ but $T \not\models \psi$.

In the next section, we use the interpretations to decompose the class $\mathcal{T}_{\text{uniq}}$ into smaller classes of causal models.

### 2.4 Classes of Models

We begin with some definitions. The *semantic graph* associated with model $T$ and $u \in \mathcal{U}$ is the directed graph $G(T, u)$ with vertex set equal to $\mathcal{X}$ and edge set $E$ such that $(X, Y) \in E$ if and only if $X$ is a non-trivial argument of $F_Y(\cdot; u)$. The semantic graph $G(T)$ associated with $T$ is the union of all graphs $G(T, u)$. Given two graphs $G, G'$, we write $G \leq G'$ when $G$ is an edge subgraph of $G'$.

**Definition 2** *A causal model $T$ is strong recursive if $G(T)$ is a DAG. It is recursive if $G(T, u)$ is a DAG for all $u \in \mathcal{U}$.*

Let $\mathcal{T}_{\text{srec}}$ and $\mathcal{T}_{\text{rec}}$ be the subclasses of models from $\mathcal{T}_{\text{uniq}}$ that are strong recursive and recursive respectively. By definition, $\mathcal{T}_{\text{srec}} \subseteq \mathcal{T}_{\text{rec}} \subseteq \mathcal{T}_{\text{uniq}}$.

We say that two models $T, T'$ are $\mathcal{L}$-equivalent, written $T \cong T'$, if they can not be distinguished by the language; i.e., $\forall \varphi [T \models \varphi \Leftrightarrow T' \models \varphi]$. We say that a class $\mathcal{T}$ is $\mathcal{L}$-contained into another class $\mathcal{T}'$, written $\mathcal{T} \leq \mathcal{T}'$, if $\forall (T \in \mathcal{T}) \exists (T' \in \mathcal{T}')[T \cong T']$. The following diagram shows the partial order induced by $\leq$ among the classes of models (modulo $\cong$). The labels at the arrows indicate whether the relation is strict or not.

$$\mathcal{T}_{\text{srec}} \xrightarrow{<} \mathcal{T}_{\text{rec}} \xrightarrow{<} \mathcal{T}_{\text{uniq}} \quad (2)$$

Not all the information in this diagram is obvious. The unstrict relations follow from the containments above noted. The fact that $\mathcal{T}_{\text{srec}} < \mathcal{T}_{\text{rec}}$ is seen with the model $T$ with $\mathcal{X} = \{X_1, X_2\}$, $\mathcal{U} = \{u, u'\}$ and equations:

$$\begin{aligned} X_1(u) &= X_2(u), & X_1(u') &= 0, \\ X_2(u) &= 0, & X_2(u') &= X_1(u') \ ; \end{aligned}$$

i.e., with graph $X_1 \rightleftarrows X_2$. The other strict relation is more difficult to see; we will prove it later.

From now on, we will only consider the equivalence classes generated by $\cong$. To simplify notation, however, we write $T$ and $\mathcal{T}$ instead of $[T]_\cong$ and $\mathcal{T}/\cong$ respectively.

## 3 Axiomatizations

An axiomatic system AX is a set of formulas (or schemata) from the language plus a set of rules of inference. Each rule of inference is a *license* that allows us to derive new formulas from previous ones. A *proof* for formula $\varphi$ from a set $\Gamma$ is a finite sequence of formulas $\varphi_1, \varphi_2, \ldots, \varphi_n = \varphi$ such that each $\varphi_i$ is either an axiom, a formula in $\Gamma$, or a formula derived using a rule of inference from previous ones. We write $\Gamma \vdash_{AX} \varphi$ if $\varphi$ can be proved from $\Gamma$ in the AX system. An axiomatic system AX is *sound* with respect to a class of models $\mathcal{T}$ iff $\Gamma \vdash_{AX} \varphi$ implies $\Gamma \models_\mathcal{T} \varphi$;[1] i.e., everything that can be proved from $\Gamma$ is valid in all models that satisfy $\Gamma$. AX is *complete* with respect to $\mathcal{T}$ iff $\Gamma \models_\mathcal{T} \varphi$ implies $\Gamma \vdash_{AX} \varphi$; i.e., everything that is valid in models that satisfy $\Gamma$ can be proved from $\Gamma$.

The following list contains all the axioms considered in this paper. It is assumed that the variables in the formulas make them valid $\mathcal{L}$-formulas; e.g., that $W$ is disjoint of $XYZ$ in A2.[2]

A1. All propositional tautologies,

A2. *Strong Union:*
$(X \twoheadrightarrow Y|Z) \Rightarrow (X \twoheadrightarrow Y|ZW)$,

---

[1] $\Gamma \models_\mathcal{T} \varphi$ means $T \models \varphi$ for all $T \in \mathcal{T}$ such that $T \models \Gamma$.

[2] Capital Roman letters will be used to denote both variables and sets of variables. If $X$ and $Y$ denote variables or sets or variables, then their union, intersection difference and complement (with respect to $\mathcal{X}$) are denoted by $XY$, $X \cap Y$, $X \setminus Y$ and $X^c$ respectively.

A3. *Left Decomposition:*
$(XW \twoheadrightarrow Y|Z) \Rightarrow (X \twoheadrightarrow Y|Z)$,

A4. *Weak Right Decomposition:*
$(X \twoheadrightarrow YW|Z)\ \&\ (X \twoheadrightarrow Y|ZW) \Rightarrow (X \twoheadrightarrow Y|Z)$,

A5. *Weak Transitivity:*
$(X \twoheadrightarrow YW|Z)\ \&\ (W \twoheadrightarrow Y|ZX) \Rightarrow (X \twoheadrightarrow Y|ZW)$,

A6. *Left Intersection:*
$(X \twoheadrightarrow Y|ZW)\ \&\ (W \twoheadrightarrow Y|ZX) \Rightarrow (XW \twoheadrightarrow Y|Z)$,

A7. *Right Intersection:*
$(X \twoheadrightarrow Y|ZW)\ \&\ (X \twoheadrightarrow W|ZY) \Rightarrow (X \twoheadrightarrow YW|Z)$,

A8. *Linearity:*
$(X \twoheadrightarrow Y|ZV)\ \&\ (X \twoheadrightarrow Y|ZU)\ \&\ (U \twoheadrightarrow V|ZXW)\ \&$
$(V \twoheadrightarrow U|ZXW) \Rightarrow (X \twoheadrightarrow Y|ZW)$,

A9. *Context Substitution:*
$(X \twoheadrightarrow YW|Z)\ \&\ (X \twoheadrightarrow V|ZYW)\ \&\ (W \twoheadrightarrow Y|ZX)$
$\Rightarrow (X \twoheadrightarrow YV|Z)$.

Axioms A2–A4, A6 and A7 appeared in Pearl [6]. The other three are new.

Let $\Gamma \subseteq \mathcal{L}$ be a set of formulas and fix an axiomatic system AX. The following definitions are standard. $\Gamma$ is a *closed* set if $\forall \varphi [\Gamma \vdash_{AX} \varphi \Rightarrow \varphi \in \Gamma]$. The *theory* associated with $\Gamma$, denoted as $\text{Th}_{AX} \Gamma$, is the smallest closed set containing $\Gamma$. $\Gamma$ is a *consistent* set if $\forall \varphi [\varphi \in \text{Th}_{AX} \Gamma \Rightarrow \neg \varphi \notin \text{Th}_{AX} \Gamma]$. A set $\Gamma' \supseteq \Gamma$ is an *extension* of $\Gamma$ if it is consistent and maximal. The theory associated with a model $T$ is the set of all formulas valid in $T$; i.e. $\text{Th}\, T = \{\varphi \in \mathcal{L} : T \models \varphi\}$.

The $\text{AX}_{\text{uniq}}$ axiomatic system is defined as the Axioms A1–A9 plus Modus Ponens (MP) as the unique rule of inference. Often we say that $\Gamma$ is $\text{AX}_{\text{uniq}}$–consistent as an abbreviation for $\Gamma$ is consistent with respect to $\text{AX}_{\text{uniq}}$. Similarly for extensions and for other axiomatic systems.

**Theorem 2** $\text{AX}_{\text{uniq}}$ *is sound with respect to* $\mathcal{T}_{\text{uniq}}$.

*Proof:* The soundness of A1 is obvious. Fix $u \in \mathcal{U}$. For clarity, we won't write the dependency in $u$. The soundness of A2, A3, A6 and A7 was proved in [2]. Axiom A4 is in [6] but its proof has not been published. Therefore, we give the proofs only for A4, A5, A8 and A9. In what follows, lower case Roman letters refer to values for the corresponding capital letters; e.g., $x, x'$ will be valuations for $X$. In each case, the suppositions are denoted from first to last by S1, S2, ...

A4. Suppose $(X \twoheadrightarrow YW|Z)$ and $(X \twoheadrightarrow Y|ZW)$, and let $V$ be disjoint from $WXYZ$. Then,

$$\begin{aligned} [x, z, v]Y &= [x, z, v, [x, z, v]W]Y & \text{(comp.)} \\ &= [x', z, v, [x, z, v]W]Y & \text{(by S2)} \\ &= [x', z, v, [x', z, v]W]Y & \text{(by S1)} \\ &= [x', z, v]Y. & \text{(comp.)} \end{aligned}$$

Thus, $(X \nrightarrow Y|Z)$.

A5. Suppose $(X \nrightarrow YW|Z)$ and $(W \nrightarrow Y|ZX)$, and let $V$ be disjoint from $XYWZ$. Then,

$$\begin{aligned}
[x,z,v,w]Y &= [x,z,v,[x,z,v]W]Y & \text{(by S2)} \\
&= [x,z,v]Y & \text{(comp.)} \\
&= [x',z,v]Y & \text{(by S1)} \\
&= [x',z,v,[x',z,v]W]Y & \text{(comp.)} \\
&= [x',z,v,w]Y. & \text{(by S2)}
\end{aligned}$$

Thus, $(X \nrightarrow Y|ZW)$.

A8. Suppose $(X \nrightarrow Y|ZV)$, $(X \nrightarrow Y|ZU)$, $(U \nrightarrow V|ZXW)$ and $(V \nrightarrow U|ZXW)$, and let $T$ be disjoint from $WXYUV$. Then,

$$\begin{aligned}
&[x,w,z,t]Y \\
&= [x,w,z,t,[x,w,z,t]U]Y & \text{(comp.)} \\
&= [x',w,z,t,[x,w,z,t]U]Y & \text{(by S2)} \\
&= [x',w,z,t,[x,w,z,t]U, & \text{(comp.)} \\
&\quad [x',w,z,t,[x,w,z,t]U]V]Y \\
&= [x',w,z,t,[x,w,z,t]U, & \text{(by S3)} \\
&\quad [x',w,z,t,[x',w,z,t]U]V]Y \\
&= [x',w,z,t,[x,w,z,t]U,[x',w,z,t]V]Y & \text{(comp.)} \\
&= [x',w,z,t,[x',w,z,t]V, & \text{(comp.)} \\
&\quad [x,w,z,t,[x,w,z,t]V]U]Y \\
&= [x',w,z,t,[x',w,z,t]V, & \text{(by S4)} \\
&\quad [x,w,z,t,[x',w,z,t]V]U]Y \\
&= [x,w,z,t,[x',w,z,t]V, & \text{(by S2)} \\
&\quad [x,w,z,t,[x',w,z,t]V]U]Y \\
&= [x,w,z,t,[x',w,z,t]V]Y & \text{(comp.)} \\
&= [x',w,z,t,[x',w,z,t]V]Y & \text{(by S1)} \\
&= [x',w,z,t]Y. & \text{(comp.)}
\end{aligned}$$

Thus, $(X \nrightarrow Y|WZ)$.

A9. Suppose $(X \nrightarrow YW|Z)$, $(W \nrightarrow Y|ZX)$ and $(X \nrightarrow V|ZYW)$. They imply, by A5, A4 and A7, $(X \nrightarrow Y|Z)$ and $(X \nrightarrow YV|ZW)$. Therefore, we only need to prove the result for $V$ and when the context $T$ is disjoint from $VXYZ$. Then,

$$\begin{aligned}
[x,z,t]V &= [x,z,t,[x,z,t]W, & \text{(comp.)} \\
&\quad [x,z,t,[x,z,t]W]Y]V \\
&= [x,z,t,[x,z,t]W,[x,z,t]Y]V & \text{(comp.)} \\
&= [x',z,t,[x',z,t]W,[x',z,t]Y]V & \text{(by S1)} \\
&= [x',z,t,[x',z,t]W, & \text{(comp.)} \\
&\quad [x',z,t,[x',z,t]W]Y]V \\
&= [x',z,t]V. & \text{(comp.)}
\end{aligned}$$

Thus, $(X \nrightarrow YV|Z)$. ∎

Some Theorems of $\mathrm{AX}_{\mathrm{uniq}}$ are:

1 *Extended Left Intersection:*
$(TX \nrightarrow Y|ZW) \& (TW \nrightarrow Y|ZX) \Rightarrow (TXW \nrightarrow Y|Z)$,

2 *Extended Right Intersection:*
$(X \nrightarrow TY|ZW) \& (X \nrightarrow TW|ZX) \Rightarrow (X \nrightarrow TYW|Z)$.

Let $\Gamma$ be a maximal consistent set of formulas with respect to $\mathrm{AX}_{\mathrm{uniq}}$. A set $P$ of variables is called the *parent* set of $Y$, also denoted as $\mathrm{Pa}(Y)$, if $((PY)^c \nrightarrow Y|P) \in \Gamma$ and $P$ is $\subseteq$–minimal. If such set $P$ does not exist, then $\mathrm{Pa}(Y)$ is defined as $Y^c$. The *syntactic graph* associated with a maximal consistent set $\Gamma$ is the directed graph $G(\Gamma)$ with vertex set equal to $\mathcal{X}$ and edge set $E$ such that $(X,Y) \in E$ if and only if $X \in \mathrm{Pa}(Y)$. From now on, if the set $\Gamma$ is clear from context, we do not write the suffix "$\in \Gamma$".

**Theorem 3** *Let $\Gamma$ be a maximal $\mathrm{AX}_{\mathrm{uniq}}$-consistent set and $T$ a causal model with exogenous domain $\mathcal{U}$. Then,*

*(i) $\mathrm{Pa}(Y)$ is unique (i.e., $G(\Gamma)$ is well defined),*

*(ii) $X \in \mathrm{Pa}(Y)$ iff $\neg(X \nrightarrow Y|W)$ for all $W$,*

*(iii) $G(T) = G(\mathrm{Th}\,T)$,*

*(iv) if $\neg(X \nrightarrow Y|Z)$, then there is a (directed) path $X \rightsquigarrow Y$ not intercepted by $Z$ in $G(\Gamma)$.*

*Proof: (i).* It is enough to show that the collection of sets that "separates" $Y$ from the rest is closed under intersections. Let $W_1, P_1$ and $W_2, P_2$ be sets of variables such that $W_1 = (P_1Y)^c$, $W_2 = (P_2Y)^c$, $(W_1 \nrightarrow Y|P_1)$ and $(W_2 \nrightarrow Y|P_2)$. Let $P = P_1 \cap P_2$. Then, by $P_1 \setminus P \subseteq W_2$, $P_2 \setminus P \subseteq W_1$ and Extended Left Intersection, $(W \nrightarrow Y|P)$ where $W = (YP)^c$.

*(ii).* Let $P = \mathrm{Pa}(Y) \setminus X$, $W = (XYP)^c$. Then, $\neg(WX \nrightarrow Y|P)$ and, by A6, $\neg(X \nrightarrow Y|WP)$. So, the only if direction follows from A2. If $\neg(X \nrightarrow Y|(XY)^c)$, then, by A2 and definition, $X \in \mathrm{Pa}(Y)$.

*(iii).* Need to prove that an edge $(X,Y) \in G(T)$ if and only if $(X,Y) \in G(\mathrm{Th}\,T)$. Observe that $\mathrm{Th}\,T$ is maximal consistent. Let $W = (XY)^c$. Then,

$$\begin{aligned}
(X,Y) &\in G(T) \\
&\iff \exists u[(X,Y) \in G(T,u)] \\
&\iff \exists u[X \text{ is non-trivial arg. of } F_Y(\,\cdot\,;u)] \\
&\iff \exists u[T(u) \vDash \neg(X \nrightarrow Y|W)] \\
&\iff \neg(X \nrightarrow Y|W) \in \mathrm{Th}\,T \\
&\iff X \in \mathrm{Pa}(Y)\,.
\end{aligned}$$

*(iv).* By reverse induction in $\#Z$ (cardinality of $Z$). Let $Z = (XY)^c$ such that $\neg(X \nrightarrow Y|Z)$. Then, $X \in \mathrm{Pa}(Y)$ so it is true. Assume it is true for all $X,Y,Z$ such that $\#Z > n$ and that $\neg(X \nrightarrow Y|Z)$ with $\#Z = n$. Then $\mathrm{Pa}(Y) \nsubseteq Z$ and, by A4 and A7, $\neg(X \nrightarrow P|ZYS)$ for some $P \in \mathrm{Pa}(Y) \setminus Z$ and $S = \mathrm{Pa}(Y) \setminus P$. Thus, by hypothesis, there is a path $X \rightsquigarrow P$ not intercepted by $ZYS$. Extend the path with the link $(P, Y)$. ∎

**Example 1:** Let $\mathcal{X} = \{X_1, \ldots, X_4\}$ and $\Gamma$ be a maximal consistent set containing:

$(X_3 X_4 \nrightarrow X_1 | \emptyset)$, $\neg(X_2 \nrightarrow X_1 | X_3 X_4)$,
$(X_1 X_3 X_4 \nrightarrow X_2 | \emptyset)$, $(X_1 \nrightarrow X_2 X_4 | \emptyset)$,
$(X_4 \nrightarrow X_3 | \emptyset)$, $\neg(X_2 \nrightarrow X_3 | X_1 X_4)$, $\neg(X_1 \nrightarrow X_3 | X_2 X_4)$,
$(X_1 X_2 \nrightarrow X_4 | X_3)$, $\neg(X_3 \nrightarrow X_4 | X_1 X_2)$, $\neg(X_1 \nrightarrow X_4 | X_2)$.

Then, $G(\Gamma)$ is the graph:

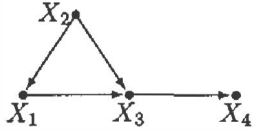

Suppose $(X_2 \nrightarrow X_4 | X_1)$, then, by $(X_1 \nrightarrow X_2 X_4 | \emptyset)$ and A5, $(X_1 \nrightarrow X_4 | X_2)$ which is a contradiction with $\neg(X_1 \nrightarrow X_4 | X_2)$. Thus, $\neg(X_2 \nrightarrow X_4 | X_1) \in \Gamma$. By Theorem 3, there must be a path $X_2 \rightsquigarrow X_4$ not intercepted by $X_1$. Fact that is clearly true in above graph. ∎

We do not know if $AX_{uniq}$ is a complete system for $\mathcal{T}_{uniq}$. The next two sections present sound and complete axiomatizations for $\mathcal{T}_{srec}$ and $\mathcal{T}_{rec}$ respectively.

### 3.1 Axiomatization for $\mathcal{T}_{srec}$

Let $\Gamma$ be a consistent set with respect to $AX_{uniq}$. We say that an $AX_{uniq}$-extension $\Gamma' \supseteq \Gamma$ is *strong-recursive* if and only if the graph $G(\Gamma')$ is a DAG. Consider the following rule of inference.

**Definition 3 (Strong Recursive Inference)**
*A formula $\varphi$ can be derived from the set $\Gamma$ by the Strong Recursive Inference (SRI) if and only if $\varphi$ holds in all strong-recursive extensions of $\Gamma$.*

Note that since the number of possible extensions is finite, checking whether a Strong Recursive Inference is valid or not is a syntactic and *decidable* criterion.

For example if $\neg(X \nrightarrow Y | \emptyset) \in \Gamma$, then (by Theorem 3) there is a path $X \rightsquigarrow Y$ in any extension of $\Gamma$. Thus, by SRI and the same Theorem, $(Y \nrightarrow X | \emptyset)$ holds in any strong-recursive extension of $\Gamma$. Thus, the SRI rule captures some notion of paths in graphs. However, it seems to be a little complex. So, do we need it? Strictly speaking, the answer is no since once $\mathcal{X}$ is fixed, then all possible SRI inferences can be encoded into a single (but very large) axiom. In the other hand, it makes things easier since we don't have to find which axiom replaces the SRI rule. We define the axiomatic system $AX_{srec}$ as the Axioms A1–A9 plus MP and SRI as the rules of inference. Our goal is to prove that it is a sound and complete system with respect to $\mathcal{T}_{srec}$.

**Theorem 4** $AX_{srec}$ *is sound with respect to* $\mathcal{T}_{srec}$.

*Proof:* Because of Theorem 2, we only need to prove that the SRI rule is sound with respect to $\mathcal{T}_{srec}$. But this is true since if $T \in \mathcal{T}_{srec}$, then $Th\,T$ is a strong-recursive extension for any $\Gamma$ such that $T \vDash \Gamma$. ∎

To prove completeness, it is enough to show that for every consistent $\Gamma$ there exists a model $T \in \mathcal{T}_{srec}$ such that $T \vDash \Gamma$. To see this, assume the claim is true and suppose that $\varphi$ is not provable from $\Gamma$. Then, $\Gamma \cup \{\neg\varphi\}$ is consistent so, by the claim, there is a model $T$ that satisfies $\Gamma$ and $\neg\varphi$; i.e., $\Gamma \nvDash_{\mathcal{T}_{srec}} \varphi$. Therefore, $\Gamma \nvdash \varphi \Rightarrow \Gamma \nvDash_{\mathcal{T}_{srec}} \varphi$ (equiv. to completeness) is true.

**Completeness**

Let $\Gamma$ be a maximal $AX_{srec}$-consistent set. We build a model for it by putting together small models whose graphs are called fragments. Each one of these models is built in order to satisfy a different negative literal $\neg(X \nrightarrow Y | Z)$ in $\Gamma$. Formally,

**Definition 4 (Fragment)** *Let $X, Y$ be two variables and $Z \subseteq \mathcal{X}$. A single-connected DAG[3] $G$ with vertex set $\mathcal{X}$ is called an $(X, Y | Z)$-fragment when for all $U, V \in G$ and $S, T \subseteq \mathcal{X}$*

- *(i) if $\neg(X \nrightarrow Y | Z)$, then $G$ has a path $X \rightsquigarrow Y$ not intercepted by $Z$,*
- *(ii) if $U \in G$, then $U$ is in $X \rightsquigarrow Y$ or $U$ is a root,*
- *(iii) if $(U \nrightarrow V | T)$, then $T$ intercepts $U \rightsquigarrow V$, and*
- *(iv) if $S$ is minimal such that $(U \nrightarrow VS | T)$, then either $T$ intercepts $U \rightsquigarrow V$, or $S$ contains an ancestor of $V$ not in $U \rightsquigarrow V$.*

**Example 2:** Intuitively, an $(X, Y | Z)$-fragment is a subgraph from $G(\Gamma)$ that witnesses the literal $\neg(X \nrightarrow Y | Z)$. Let's consider the literal $\neg(X_1 \nrightarrow X_4 | \emptyset)$ in Example 1. We show that the following graph is an $(X_1, X_4 | \emptyset)$-fragment:

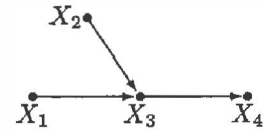

Clearly, it satisfies *(i)* and *(ii)* of Definition 4. For *(iii)*, note that the only relevances that need to be considered are $(X_1 \nrightarrow X_4 | X_3)$ and $(X_2 \nrightarrow X_4 | X_3)$ but $X_3$ intercepts the paths $X_1 \rightsquigarrow X_4$ and $X_2 \rightsquigarrow X_4$. For *(iv)*, need to consider $(X_1 \nrightarrow X_2 X_4 | \emptyset)$. Since $X_2$ is an ancestor of $X_4$ not in $X_1 \rightsquigarrow X_4$ then *(iv)* is true. Therefore, the graph is a $(X_1, X_4 | \emptyset)$-fragment. ∎

---

[3]That is, the undirected graph associated with $G$ has no cycles.

**Theorem 5** *Let $\Gamma$ be a maximal consistent set with respect to $\text{AX}_{\text{srec}}$. Then, $G(\Gamma)$ contains a $(X, Y|Z)$-fragment for all variables $X, Y \in \mathcal{X}$ and set $Z \subseteq \mathcal{X}$.*

**Lemma 6** *If $\neg(X \not\to Y|ZW)$ and $\neg(X \not\to YW|Z)$, then there is an $(X, Y|ZW)$-fragment where $W$ has no $Y$ ancestors.*

The proofs of Theorem 5 and Lemma 6 are in the full paper [1]. The next theorem is the main result of this section.

**Theorem 7** $\text{AX}_{\text{srec}}$ *is complete with respect to $\mathcal{T}_{\text{srec}}$.*

*Proof:* Let $\Gamma$ be maximal $\text{AX}_{\text{srec}}$-consistent. Let $U, V \in \mathcal{X}$ and maximal $T \subseteq \mathcal{X}$ such that $\neg(U \not\to V|T)$. Let $G \leq G(\Gamma)$ be the $(U, V|T)$-fragment given by Theorem 5. Define the recursive model $T_G$ with singleton $\mathcal{U}_G$ and $\mathcal{D}_X$ equal to the non-negative integers for all $X \in \mathcal{X}$, by[4]

$$Y = \begin{cases} 0 & \text{if there is } P \text{ s.t.} \\ & P \in \text{Pa}_G(Y), P = 0, \\ \max\{X : X \in \text{Pa}_G(Y)\} & \text{otherwise.} \end{cases}$$

This equation depends on the fixed $G$ so we write $Y^G$ when this dependency needs to be shown. By construction and properties (i)-(iii) of fragments, $T_G \models \neg(U \not\to V|Z)$ and $T_G \models (\tilde{U} \not\to \tilde{V}|\tilde{T})$ if $(\tilde{U} \not\to \tilde{V}|\tilde{T})$.

Let $\mathcal{U}$ be the collection of all fragments, and $T$ the causal model with domain $\mathcal{U}$ and equations $Y(\cdot; G) = Y^G(\cdot)$ for $G \in \mathcal{U}$. In the rest, we show using induction that $T$ satisfies all literals in $\Gamma$; i.e., $T \models \Gamma$.

First consider all literals whose first and second components are single variables. Let $X, Y \in \mathcal{X}$ and $Z \subseteq \mathcal{X}$. If $\neg(X \not\to Y|Z)$, then there is an $(X, Y|Z)$-fragment $G$ such that $X \rightsquigarrow Y$ isn't intercepted by $Z$. By construction, $T_G \models \neg(X \not\to Y|Z)$ which implies $T \models \neg(X \not\to Y|Z)$. Assume now $(X \not\to Y|Z)$ and let $G$ be any $(U, V|T)$-fragment. By definition, $Z$ intercepts all paths $X \rightsquigarrow Y$ in $G$, so by construction $T_G \models (X \not\to Y|Z)$. Thus, $T \models (X \not\to Y|Z)$.

The next step is to show that $T$ satisfies all literals whose first component is a single variable. Let $\neg(X \not\to YW|Z)$ be a such negative literal with $Y \in \mathcal{X}, W \subseteq \mathcal{X}$. Without loss of generality assume that $\neg(X \not\to Y|ZW)$. By Lemma 6, there is a fragment $G$ such that $ZW$ doesn't intercepts $X \rightsquigarrow Y$ and $W$ contains no ancestor of $Y$. By construction, $T_G \models \neg(X \not\to YW|Z)$. So, $T \models \neg(X \not\to YW|Z)$. Assume now $(X \not\to YW|Z)$ and, without loss of generality that $\neg(X \not\to Y|Z)$ since otherwise $T \models (X \not\to YW|Z)$ (the reader can check it using A4). Let $G$ be an $(U, V|T)$-fragment such that $Z$ doesn't intercept path $X \rightsquigarrow Y$. Then, (iv)

---

[4] An empty max is defined to be 0.

of Definition 4 and construction imply $T_G \models (X \not\to YW|Z)$. Thus, $T \models (X \not\to YW|Z)$.

The last step is for general literals. Note that A3 and A6 imply $(XW \not\to Y|Z) \Leftrightarrow (X \not\to Y|ZW) \& (W \not\to Y|ZX)$. Therefore, by above cases, $(XW \not\to Y|Z)$ if and only if $T \models (XW \not\to Y|Z)$. ∎

**Example 3:** Consider the $(X_1, X_4|\emptyset)$-fragment $G$ in Example 2. The construction of Theorem 7 gives the model $T_G$:

$$X_1 = 0,$$
$$X_2 = 0,$$
$$X_4 = X_3,$$
$$X_3 = \begin{cases} 0 & \text{if } X_1 = 0 \text{ or } X_2 = 0, \\ \max\{X_1, X_2\} & \text{otherwise.} \end{cases}$$

We now prove that $T_G$ satisfies $\neg(X_1 \not\to X_4|\emptyset)$ and all positive literals in Example 1. It is easy to check that $[X_1 = 0, X_2 = 1]X_4 = 0$ and $[X_1 = 1, X_2 = 1]X_4 = 1$. Thus, $\neg(X_1 \not\to X_4|X_2)$ which implies $\neg(X_1 \not\to X_4|\emptyset)$. Clearly, the model satisfies $(X_4 \not\to X_3|\emptyset)$, $(X_1X_3X_4 \not\to X_2|\emptyset)$ and $(X_2X_3X_4 \not\to X_1|\emptyset)$. It also satisfies $(X_1X_2 \not\to X_4|X_3)$ since the equation for $X_4$ depends only in $X_3$. We need to show that it satisfies $(X_1 \not\to X_2X_4|\emptyset)$;

$$[x_1]X_4 = [x_1, [x_1]X_2]X_4$$
$$= [x_1, X_2 = 0]X_4$$
$$= [x_1, X_2 = 0, [x_1, X_2 = 0]X_3]X_4$$
$$= [x_1, X_2 = 0, X_3 = 0]X_4$$
$$= 0,$$

independent of $x_1$. Thus, $T_G \models (X_1 \not\to X_2X_4|\emptyset)$. ∎

### 3.2 Axiomatization for $\mathcal{T}_{\text{rec}}$

Let $\Gamma$ be a consistent set with respect to $\text{AX}_{\text{uniq}}$. We say that an $\text{AX}_{\text{uniq}}$-extension $\Gamma' \supseteq \Gamma$ is *recursive* if and only if the graph $G(\Gamma')$ contains an $(X, Y|Z)$-fragment for all $X, Y \in \mathcal{X}$ and $Z \subseteq \mathcal{X}$. Similar to SRI, we make

**Definition 5 (Recursive Inference)** *A formula $\varphi$ can be derived from set $\Gamma$ by Recursive Inference (RI) if and only if $\varphi$ holds in all recursive extensions of $\Gamma$.*

As with SRI, all RI inferences are decidable. Let $\text{AX}_{\text{rec}}$ be the axiomatic systems defined as the Axioms A1–A9 plus MP and RI as the rules of inference. Our goal is to prove that it is sound and complete for $\mathcal{T}_{\text{rec}}$.

**Theorem 8** $\text{AX}_{\text{rec}}$ *is sound with respect to $\mathcal{T}_{\text{rec}}$.*

*Proof:* Because of Theorem 2, we only need to prove that the RI rule is sound. Let $T \in \mathcal{T}_{\text{rec}}$ and $\Gamma$ such that $T \models \Gamma$. Let $X, Y \in \mathcal{X}$ and $Z \subseteq \mathcal{X}$. If $(X \not\twoheadrightarrow Y|Z)$, then the empty graph is a $(X,Y|Z)$-fragment. If $\neg(X \not\twoheadrightarrow Y|Z)$, then there is $u \in \mathcal{U}$ such that $T(u) \vDash \neg(X \not\twoheadrightarrow Y|Z)$. By Theorem 5, $G(\operatorname{Th} T(u))$ contains a $(X,Y|Z)$-fragment. Then, Th $T$ is a recursive extension of $\Gamma$. ∎

Let $T_1$ and $T_2$ be two functional causal models with exogenous domains $\mathcal{U}_1, \mathcal{U}_2$, and equations $F_Y^i$ for $i = 1, 2$ and $Y \in \mathcal{X}$. The direct sum of $T_1$ and $T_2$, denoted by $T_1 \oplus T_2$, is defined as the causal model $T$ with exogenous domain $\mathcal{U} = \{1\} \times \mathcal{U}_1 \cup \{2\} \times \mathcal{U}_2$ and equations given by $F_Y(\cdot ; (i, u)) \stackrel{\text{def}}{=} F_Y^i(\cdot ; u)$ for $i = 1, 2$. Abusing notation, we say that the causal model $T_1 \oplus \cdots \oplus T_n$ is the model with exogenous domain $\mathcal{U} = \bigcup_i \{i\} \times \mathcal{U}_i$ and equations given by $F_Y(\cdot ; (i, u)) = F_Y^i(\cdot ; u)$ for $1 \leq i \leq n$.

**Theorem 9** *Let $T, T_1, \ldots, T_n \in \mathcal{T}_{\text{uniq}}$ be causal models. Then,*

*(i) $T \cong T_1 \oplus \cdots \oplus T_n$ if and only if $\operatorname{Th} T$ is the unique $\operatorname{AX}_{\text{uniq}}$-extension of*

$$\bigcup_{i=1}^n \operatorname{Lit}^-(\operatorname{Th} T_i) \ \cup \ \bigcap_{i=1}^n \operatorname{Lit}^+(\operatorname{Th} T_i) . \quad (3)$$

*(ii) If $T \cong T_1 \oplus \cdots \oplus T_n$, then $G(T) = G(T_1) \cup \cdots \cup G(T_n)$.*

*Proof:* *(i).* Denote with $\Gamma$ the set of formulas in Eq.(3). First we prove that $\Gamma$ has only one extension. Let $\varphi \in \mathcal{L}$ be a positive literal. If $\exists i[T_i \nvDash \varphi]$, then $\exists i[T_i \vDash \neg \varphi]$ which implies $\neg \varphi \in \Gamma$. If $\forall i[T_i \vDash \varphi]$, then $\varphi \in \cap_{i=1}^n \operatorname{Lit}^+(\operatorname{Th} T_i)$ which implies $\varphi \in \Gamma$. For negative literals consider $\psi = \neg \varphi$. Thus, $\Gamma$ is a maximal consistent set, so its unique extension is $\operatorname{Th}_{\text{uniq}} \Gamma$.

Assume $T \cong T_1 \oplus \cdots \oplus T_n$, and $\varphi$ a positive literal. If $T \vDash \varphi$, then $\forall (i,u)[T(i,u) \vDash \varphi]$ which implies $\forall i[\varphi \in \operatorname{Lit}^+(\operatorname{Th} T_i)]$, so $\varphi \in \Gamma$. If $T \nvDash \varphi$, then $\exists (i,u)[T(i,u) \nvDash \varphi]$ which implies $\exists i[\varphi \notin \operatorname{Lit}^+(\operatorname{Th} T_i)]$, so $\varphi \notin \Gamma$. Therefore, $\operatorname{Lit}(\operatorname{Th} T) = \operatorname{Lit} \Gamma$ which implies the result.

Now, assume $\operatorname{Th} T = \operatorname{Th}_{\text{uniq}} \Gamma$. Let $\varphi$ be a positive literal. Then, $\varphi \in \operatorname{Th} T$ iff

$$\varphi \in \bigcap_{i=1}^n \operatorname{Lit}^+(\operatorname{Th} T_i) \iff \varphi \in \operatorname{Th}(T_1 \oplus \cdots \oplus T_n).$$

*(ii).* Assume $T \cong T_1 \oplus \cdots \oplus T_n$. Let $X, Y$ be two variables and $W = (XY)^c$. Then, $(X,Y) \in G(T)$ iff $(X,Y) \in G(\operatorname{Th}(T_1 \oplus \cdots \oplus T_n))$ iff $\exists i[T_i \vDash \neg(X \not\twoheadrightarrow Y|W)]$ iff $\exists i[(X,Y) \in G(T_i)]$. ∎

**Completeness**

To motivate our proof for completeness, observe that a model $T \in \mathcal{T}_{\text{rec}}$ is composed of different $\mathcal{T}_{\text{srec}}$ models (one per $u \in \mathcal{U}$), and that each negative literal satisfied by $T$ is satisfied by one of those models. We will do something similar but with sets of formulas. Thus, to each maximal consistent set $\Gamma$ and negative literal $\varphi \in \Gamma$, we assign a maximal $\operatorname{AX}_{\text{srec}}$-consistent set $\Gamma' \subseteq \Gamma$ containing $\varphi$. All such sets are then use to build $\mathcal{T}_{\text{srec}}$ models that combine into a model $T \in \mathcal{T}_{\text{rec}}$ such that $T \vDash \Gamma$.

Let $\psi_0, \psi_1, \ldots, \psi_n$ be a fixed enumeration of all literals in $\mathcal{L}$. Fix a maximal $\operatorname{AX}_{\text{rec}}$-consistent set $\Gamma$. Let $X, Y$ be variables, $\varphi = \neg(X \not\twoheadrightarrow Y|Z)$ a negative literal in $\Gamma$, and define the sequence of sets $\{\Psi_i\}_{i=1}^n$ as

$$\Psi_0 = \operatorname{Th}_{\text{srec}}(\{\varphi\} \cup \operatorname{Lit}^+ \Gamma)$$

$$\Psi_{i+1} = \begin{cases} \operatorname{Th}_{\text{srec}}(\Psi_i \cup \{\psi_i\}) & \text{if } \Psi_i \cup \{\psi_i\} \text{ is} \\ & \operatorname{AX}_{\text{srec}}\text{-consistent,} \\ \operatorname{Th}_{\text{srec}}(\Psi_i \cup \{\neg \psi_i\}) & \text{otherwise.} \end{cases}$$

We want to prove that each $\Psi_i$ is $\operatorname{AX}_{\text{srec}}$-consistent. By construction, it is enough to show that $\Psi_0$ is consistent. Fact that is true since $\Gamma$ is $\operatorname{AX}_{\text{rec}}$-consistent; i.e., $G(\Gamma)$ contains an $(X,Y|Z)$-fragment for $\varphi$. Note that $\Psi_n$ is maximal and that $\operatorname{Lit}^- \Psi_i \subseteq \operatorname{Lit}^- \Gamma$ since $\operatorname{Lit}^+ \Gamma \subseteq \Psi_0$. We call $\Psi_n$ a $(\varphi, \Gamma)$-*foliation*.

**Theorem 10** $\operatorname{AX}_{\text{rec}}$ *is complete with respect to $\mathcal{T}_{\text{rec}}$.*

*Proof:* Let $\Gamma$ be a maximal $\operatorname{AX}_{\text{rec}}$-consistent, and $\Psi$ a $(\varphi, \Gamma)$-foliation. Denote with $T_\Psi$ the strong recursive model such that $T_\Psi \vDash \Psi$ (it exists by completeness of $\operatorname{AX}_{\text{srec}}$). By a simple induction in the structure of negative literals (similar to that in proof of Theorem 7), it can be shown that for all literal $\neg(U \not\twoheadrightarrow Z|T)$ in $\Gamma$, there exists a foliation $\Psi$ such that $T_\Psi \vDash \neg(U \not\twoheadrightarrow Z|T)$. Let $\mathcal{F} = \{ \Psi : \Psi \text{ is a } (\varphi, \Gamma)\text{-foliation} \}$, and $T = \bigoplus_{\Psi \in \mathcal{F}} T_\Psi$. Then, by Theorem 9, $T \vDash \Gamma$. ∎

Now, we prove the claim made in diagram (2) about the subclasses of causal models.

**Theorem 11** $\mathcal{T}_{\text{rec}} < \mathcal{T}_{\text{uniq}}$.

*Proof:* Consider the model $T$ with variables $\{X_1, \ldots, X_4\}$, singleton $\mathcal{U}$ and equations

$$X_1 = 1, \quad X_2 = (X_1, X_3), \quad X_3 = X_2\!\downarrow_1, \quad X_4 = X_2\!\downarrow_2 .$$

It is a non-recursive model that satisfies $\neg(X_1 \not\twoheadrightarrow X_4|\emptyset)$ and $(X_1 \not\twoheadrightarrow X_4|Z)$ for $Z \in \{X_2, X_3\}$. Note that $G(T) = G(\operatorname{Th} T)$ is the graph with edges $\{X_1 \to X_2 \to X_4, X_2 \rightleftarrows X_3\}$, and that $T$ satisfies $\neg(X_2 \not\twoheadrightarrow X_4|X_1X_3)$, $\neg(X_1 \not\twoheadrightarrow X_3|X_4)$ and $\neg(X_2 \not\twoheadrightarrow X_3|X_4)$. It can be shown that $\operatorname{Th} T$ has no recursive extension, so it is not $\operatorname{AX}_{\text{rec}}$-consistent. Therefore, there is no recursive model $\mathcal{L}$-equivalent to it. ∎

## 4 Applications

In this section we consider the question of how to use information about causal relevance for the problem of identifying causal structure (also known as the problem of discovering causal structure). Fix a model $T$ and let $\Gamma$ be a set of causal relevances that hold in $T$; i.e., $T \models \Gamma$. If $T \in \mathcal{T}_{\text{srec}}$, then the completeness of $AX_{\text{srec}}$ and Theorem 3 imply $G(T)$ is equal to $G(\Gamma')$ for some $AX_{\text{srec}}$-extension $\Gamma'$ of $\Gamma$. Thus, the graph defined as

$$\mathcal{G}(\Gamma) \stackrel{\text{def}}{=} \bigcap_{\Gamma'} G(\Gamma'),$$

where the intersection is over all $AX_{\text{srec}}$-extensions of $\Gamma$, is an edge subgraph of $G(T)$. Completeness is needed here in order to guarantee that every $\Gamma'$ has a corresponding model in $\mathcal{T}_{\text{srec}}$. Suppose further that we can buy one of several possible information options $\Gamma_i \supseteq \Gamma$. The problem is to decide which is the best option. Let $c_i$ be the cost associated to $\Gamma_i$. A rational approach would be to buy the set that minimizes the pondered cost $C_i \stackrel{\text{def}}{=} c_i / \#(\mathcal{G}(\Gamma_i) - \mathcal{G}(\Gamma))$ where $\#G$ denotes the number of edges in graph $G$. The formula for $C_i$ considers that the benefit of the information is proportional to the number of new edges provided by it. Note that if $\Gamma_i$ generates no new edges, then $C_i = \infty$. Similar definitions can be made when $T \in \mathcal{T}_{\text{rec}}$. Observe that we do not claim this can be implemented in an efficient way. Nonetheless, this calculus can be integrated into other methods of structure identification by using it to prune the search space; e.g., once we know $\neg(X \not\rightarrow Y|Z)$, then the graph we are looking for must has a path $X \rightsquigarrow Y$ not intercepted by $Z$.

Another application of the calculus is for testing whether a model is recursive or not. Since $AX_{\text{rec}}$ is complete with respect to $\mathcal{T}_{\text{rec}}$, then a *necessary* test for recursiveness is $AX_{\text{rec}}$-consistency. Thus, if $\Gamma$ is not $AX_{\text{rec}}$-consistent, then $\Gamma$ corresponds to a non-recursive model. An argument like this one was used in the proof of Theorem 11.

## 5 Summary

There are three main contributions in this paper. First, it gives a finer decomposition $\mathcal{T}_{\text{uniq}}$ into the two subclasses $\mathcal{T}_{\text{srec}}$ and $\mathcal{T}_{\text{rec}}$. This decomposition is necessary for the formalization of causal relevance from the three-place relation $(X \not\rightarrow Y|Z)$, and it is useful as shown in the applications of the calculus. The second contribution is the collection of axiomatizations $AX_{\text{uniq}}$, $AX_{\text{rec}}$ and $AX_{\text{srec}}$. $AX_{\text{uniq}}$ extends the system in [6] with the three new axioms A5, A8 and A9. It is a sound axiomatization with respect to the class $\mathcal{T}_{\text{uniq}}$. $AX_{\text{rec}}$ is a stronger version of $AX_{\text{uniq}}$ that is sound and complete with respect to $\mathcal{T}_{\text{rec}}$. It is equal to $AX_{\text{uniq}}$ plus the rule RI. The axiomatization $AX_{\text{srec}}$, a stronger version $AX_{\text{rec}}$, is sound and complete with respect to $\mathcal{T}_{\text{srec}}$. It is equal to $AX_{\text{uniq}}$ plus the rule SRI. Finally, we show how the calculus of causal relevance can be used to identify causal structure. This is a non-probabilistic approach that can be used to infer structure from non-observational data. All other methods we know of are based on probabilistic inference (e.g., [9]). It would be interesting to study the similarities and difference between these approaches.

Some open questions are if $AX_{\text{uniq}}$ is a complete system with respect to $\mathcal{T}_{\text{uniq}}$, if the collection A1–A9 is an independent set of Axioms, and which Axioms can be used to replace the RI and SRI rules.

## Acknowledgments

Thanks to Carlos Brito and Judea Pearl for their useful comments in a first version of this paper. They spotted a number of errors, omissions and ways to improve it. Thanks also Jin Tian and people at the AI seminar at UCLA.